\documentclass[10pt, letterpaper, oneside]{article}
\usepackage[margin=25mm]{geometry}
\usepackage[utf8]{inputenc}
\usepackage{array}
\usepackage{tabu}
\usepackage{longtable}
\usepackage{tabularx}
\usepackage{tabulary}
\usepackage{graphicx}
\usepackage{siunitx}
\usepackage{seqsplit}
\usepackage{float}
\usepackage{booktabs}
\usepackage{natbib}
\setcitestyle{numbers}
\usepackage{caption}
\usepackage{amssymb}
\usepackage[affil-it]{authblk}

\providecommand{\keywords}[1]
{
  \small	
  \textbf{\textit{Keywords---}} #1
}

\title{All Data Inclusive, Deep Learning Models to Predict Critical Events in the Medical Information Mart for Intensive Care III Database (MIMIC III)}
\date{\vspace{-5ex}}

\author[1]{Anubhav Reddy Nallabasannagaari}
\author[1]{Madhu Reddiboina}
\author[2]{Ryan Seltzer}
\author[1]{Trevor Zeffiro}
\author[3]{Ajay Sharma}
\author[4]{Mahendra Bhandari \thanks{Corresponding author}\thanks{E-mail: mahendra@vattikutifoundation.com}}
\affil[1]{Rediminds, Southfield, Michigan, USA}
\affil[2]{Translational Analytics and Statistics, Tucson, Arizona, USA}
\affil[3]{University of Liverpool, Liverpool, UK}
\affil[4]{Vattikuti Urology Institute, Detroit, Michigan, USA}

\begin{document}

\maketitle

\begin{abstract}
\noindent \textbf{Purpose:}
Intensive care clinicians need reliable clinical practice tools to preempt unexpected critical events that might harm their patients in intensive care units (ICU), to pre-plan timely interventions, and to keep the patient’s family well informed. The conventional statistical models are built by curating only a limited number of key variables, which means a vast unknown amount of potentially precious data remains unused.  Deep learning models (DLMs) can be leveraged to learn from large complex datasets and construct predictive clinical tools to develop evidence-based management protocols. \\ \\
\textbf{Materials and Methods:}
This retrospective study was performed using 42,818 hospital admissions involving 35,348 patients, which is a subset of the MIMIC-III dataset. Natural language processing (NLP) techniques were applied to build DLMs to predict in-hospital mortality (IHM) and length of stay $\geq$7 days (LOS).\\ \\
\textbf{Results:}
Over 75 million events across multiple data sources were processed, resulting in over 355 million tokens. DLMs for predicting IHM using data from all sources (AS) and chart data (CS) achieved an AUC-ROC of 0.9178 and 0.9029, respectively, and PR-AUC of 0.6251 and 0.5701, respectively. DLMs for predicting LOS using AS and CS achieved an AUC-ROC of 0.8806 and 0.8642, respectively, and PR-AUC of 0.6821 and 0.6575, respectively. The observed AUC-ROC difference between models was found to be significant for both IHM and LOS at p=0.05. The observed PR-AUC difference between the models was found to be significant for IHM and statistically insignificant for LOS at p=0.05.\\ \\
\textbf{Conclusions:} 
In this study, deep learning models were constructed using data combined from a variety of sources in Electronic Health Records (EHRs) such as chart data, input and output events, laboratory values, microbiology events, procedures, notes, and prescriptions. It is possible to predict in-hospital mortality with much better confidence and higher reliability from models built using all sources of data.
\end{abstract}


\section{Introduction}
Healthcare, one of the fastest-growing segments of the digital universe, is expected to reach 2,314 exabytes of data by 2020.\cite{Www.emc.com} Effective utilization of this data would go a long way in the better prediction of the course of illness, thereby making the treatment actions increasingly predictable, safe, and consistent. Electronic health records (EHRs) are a digital form of comprehensive patient data reservoir comprising vital sign measurements, intake output values, medications, interventions, laboratory tests, imaging reports, and caregiver notes. Accurate deep learning models (DLMs) based on the entire EHRs are expected to empower the clinicians to tailor the treatment protocols specific for a patient and to facilitate both internal and external quality control.\cite{Chawla2013} For instance, Berlyand et al. reported how machine learning algorithms could quickly assess patients presenting to an emergency room for efficient triage, optimal resource utilization, and guidance in the timely management of sepsis and cardiac dysfunction before a cascade of events can lead to irreversible damage.\cite{berlyand2018artificial} 

Predicting in-hospital mortality and length of stay for patients in ICU has been an area of active research. Timely and accurate predictions of these clinical outcomes can empower clinicians to assess the gravity of a patient’s condition, provide a window of opportunity for interventions, and contribute towards cost-effective management of hospital resources. Despite this ground-breaking potential, much of the data available in digital healthcare records is not widely used for building predictive models. Current Intensive Care Unit (ICU) scoring systems such as Acute Physiology, Age, Chronic Health Evaluation (APACHE), and Simple Acute Physiology Score (SAPS) were developed using a curated list of variables. These ICU scoring systems did not account for caregiver’s notes in EHRs. The caregiver notes can store subtle aspects of a patient’s disease state, not-so-obvious features of medical history, imperceptible treatment interactions, and unnoticed natural history of illness. Unfortunately, these notes are unstructured and not in a readily analyzable format. 

Application of Natural Language Processing (NLP) methodologies on caregiver notes have been an area of active research.\cite{Shickel2018,Zeng2019} Leveraging NLP techniques can enable us to overcome the challenges of feature selection and make it possible to utilize, and merge structured and unstructured data to build DLMs with high predictive value. Technical solutions that leverage data, structured and unstructured, are expected to enhance the performance of predictive models because they would include meaningful and hidden insights.\cite{Krumholz2014} In recent years the neural networks have been widely applied in NLP tasks such as language translation, abstract summarization, and sentiment analysis.\cite{hirschberg2015advances}

We hypothesize that DLMs built using complete patient data can predict IHM and LOS with higher performance than models built using a subset of patient data. In order to build models that would learn from the entirety of EHR data, NLP methodologies and deep learning neural networks were leveraged in this study. DLMs were constructed to predict in-hospital mortality (IHM) and length of stay $\geq$7 days (LOS) on MIMIC-III v1.4 (Medical Information Mart for Intensive Care III) dataset. MIMIC-III is a large single-center database comprising information related to patients admitted to critical care units of Beth Israel Deaconess Medical Center Boston, Massachusetts, a large tertiary care hospital.\cite{Johnson2016} For each outcome, a model built using only Chart Events was compared against a model developed using multiple sources: Chart Events, Input Events, Output Events, Lab Events, Microbiology Events, Procedure Events, Note Events, and Prescriptions.

This paper is divided into five sections. Section 1 introduces the context, objective, background, and related work of the study. Section 2 presents the material and methods used to complete the study. Section 3 presents the observed results. Section 4 discusses the observed results in the context of the objective and related work. Section 5 concludes and summarizes the study.   

\subsection{Background and Related Work}
Recent efforts have been made to improve upon current ICU scoring systems using machine learning and deep learning. Harutyunyan et al. constructed linear and neural models for four clinical tasks and evaluated the effect of deep supervision, multitask training, and data-specific changes on the performance of these models. They reported that LSTM models outperformed linear models in their study across all tasks.\cite{2017arXiv170307771H} Davoodi et al. proposed a Deep Rule-Based Fuzzy System (DRBFS) for predicting in-hospital mortality, where they investigated rule generation using clustering on mixed attributes.\cite{DAVOODI201848} Yu et al. constructed a multi-task recurrent neural network with an attention mechanism to predict in-hospital mortality and observed that it outperformed the Simplified Acute Physiology Score (SAPS-II).\cite{yu2019using} Boag et al. explored different methodologies to represent clinical notes and evaluated their performance for predicting multiple clinical outcomes.\cite{boag2018s} Si et al. constructed a multi-task convolutional network on clinical notes to predict multiple mortality tasks and observed that multi-task prediction enabled small gains on single task prediction.\cite{si2019deep} In these studies, the authors' efforts primarily focused on only a few curated variables or clinical notes. Some recent efforts were made to combine structured data and unstructured clinical notes. It was observed that combining clinical notes and physiologic variables improved the performance of the models.\cite{jin2018improving,weissman2018inclusion}

Selecting and curating variables from large datasets such as EHRs, which can potentially contain thousands of variables, can be a pervasive and laborious task.\cite{press_2016} Also, this results in discarding a large amount of vital patient data, potentially valuable hidden insights, available in EHRs.  Goldstein et al. in a systemic review of predictive models, predominantly traditional linear models on heart failure, reported that all the models were built with data collected from electronic health records using a limited number of variables (a median of 27 variables).\cite{Goldstein2017} In a recent study, NLP was used to combine multiple FHIR resources to develop deep learning models. It was observed that the deep learning models significantly outperformed the augmented Early Warning Score (aEWS), which was a 28-factor logistic regression model.\cite{Rajkomar2018} While using comprehensive EHR data is recognized as essential, the difficulty in making it usable remains a stout obstacle.   
\section{Material and Methods}
The research methodology used in this study comprises the following five key steps: 
\begin{enumerate}
    \item Understanding the current paradigm for using EHRs to construct deep learning models for predicting In-hospital mortality and length of stay for ICU patients (Section 1 and subsection 1.1).
    \item Exploring and understanding the data. Subsection 2.1 introduces the material, data selection process, and data characteristics. 
    \item Preparing data for training. Subsection 2.2 describes the process used to prepare and transform the data for training and evaluation.
    \item Constructing and training deep learning models. Subsection 2.3 describes the model architectures, and subsection 2.4 discusses model training and experiments conducted to determine hyperparameter settings. 
    \item Evaluating model performance using the test dataset. Subsection 2.5 describes the model evaluation and statistical analysis process. Section 3 presents the results observed during training and evaluation.
\end{enumerate}

\subsection{Dataset}
\begin{figure}[H]
	\centering
		\caption{A breakdown of the data selection process. All percentages are reported against total admissions.}
		\label{fig:dataselection}
		\includegraphics[width=0.85\textwidth]{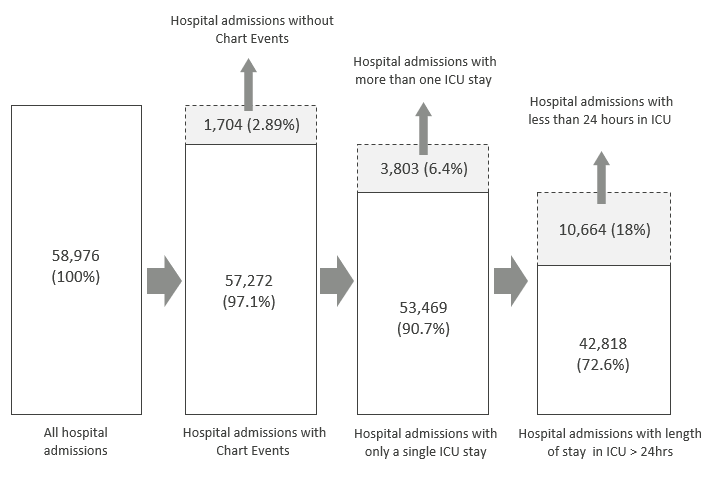}
\end{figure}

\begin{figure}[H]
	\centering
		\caption{An illustration of total available events versus events used for training across each data source.}
		\label{fig:events_24hrs}
		\includegraphics[width=0.85\textwidth]{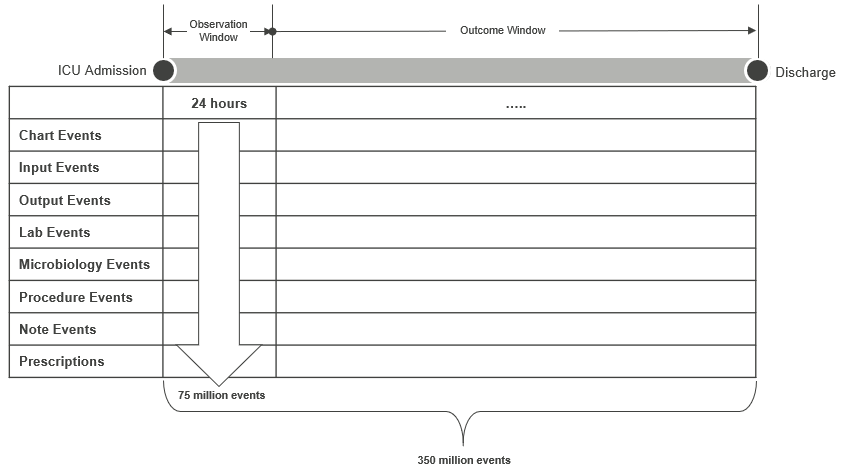}
\end{figure}

The MIMIC-III database comprises deidentified health-related data of over 40,000 patients who were admitted to ICU at Beth Israel Deaconess Medical Center between 2001-2002. The dataset was acquired by following the instruction presented on https://mimic.physionet.org/. It contains information such as vital signs, medications, laboratory measurements, observations and caregiver notes, fluid intake and output records, procedure codes, diagnostic codes, hospital length of stay, and survival data.\cite{Johnson2016} Our study was limited to 42,818 hospital admission episodes of 35,348 patients. Every hospital admission comprised of only a single, at least 24 hours long ICU stay (Figure \ref{fig:dataselection}). MIMIC-III dataset contained over 350 million events comprising multiple data points across various sources of data. We extracted 75 million events across all data sources from the first 24 hours in ICU (Figure \ref{fig:events_24hrs}). Table \ref{tab:table_datasources} describes the list of data sources, their description, MIMIC-III tables used for creating the data sources, and summary of events processed during this study. Other tables were excluded from this study as they either contained redundant data, administrative data, or billing information.

\begin{longtable}[H]{|l|m{2cm}|l|r|r|}
\caption{Description and summary of the number of events across all data sources.MIMIC Tables column below shows the table names as available in MIMIC-III v1.4 dataset used to create each data source. Only events within the first 24 hours of ICU stay were used in this study.}
\label{tab:table_datasources}\\
\hline
Data Sources & Description & MIMIC Tables & \multicolumn{1}{l|}{\begin{tabular}[c]{@{}l@{}}Total Events \\ (Millions)/\\ (\% of total)\end{tabular}} & \multicolumn{1}{l|}{\begin{tabular}[c]{@{}l@{}}Used Events\\ (Millions)/\\ (\% of total)\end{tabular}} \\ \hline
\endfirsthead
\multicolumn{5}{c}%
{{\bfseries Table \thetable\ continued from previous page}} \\
\hline
Data Sources & Description & MIMIC Tables & \multicolumn{1}{l|}{\begin{tabular}[c]{@{}l@{}}Total Events \\ (Millions)/\\ (\% of total)\end{tabular}} & \multicolumn{1}{l|}{\begin{tabular}[c]{@{}l@{}}Used Events\\ (Millions)/\\ (\% of total)\end{tabular}} \\ \hline
\endhead
\begin{tabular}[c]{@{}l@{}}Chart\\ Events\end{tabular} & Charted data available for a patient. & \begin{tabular}[c]{@{}l@{}}chartevents\\ admissions\\ icustays\\ d\_items\end{tabular} & 330.0 (85.73) & 61.0 (80.69) \\ \hline
\begin{tabular}[c]{@{}l@{}}Input\\ Events\end{tabular} & Fluids administered to the patient. & \begin{tabular}[c]{@{}l@{}}inputevents\_cv\\ inputevents\_mv\\ d\_items\end{tabular} & 20.0 (5.19) & 3.6 (4.76) \\ \hline
\begin{tabular}[c]{@{}l@{}}Output\\ Events\end{tabular} & Fluids that have been excreted by or extracted from the patient. & \begin{tabular}[c]{@{}l@{}}outputevents\\ d\_items\end{tabular} & 4.0 (1.03) & 0.9 (1.19) \\ \hline
\begin{tabular}[c]{@{}l@{}}Lab \\ Events\end{tabular} & Laboratory-based measurements. & \begin{tabular}[c]{@{}l@{}}labevents\\ d\_labitems\end{tabular} & 24.0 (6.24) & 7.4 (9.79) \\ \hline
\begin{tabular}[c]{@{}l@{}}Microbiology\\ Events\end{tabular} & Microbiology information, including tests performed and sensitivities. & microbiologyevents & 0.6 (0.16) & 0.2 (0.26) \\ \hline
\begin{tabular}[c]{@{}l@{}}Procedure\\ Events\end{tabular} & Procedures for patients. & \begin{tabular}[c]{@{}l@{}}procedureevents\_mv\\ d\_items\end{tabular} & 0.3 (0.08) & 0.1 (0.13) \\ \hline
\begin{tabular}[c]{@{}l@{}}Note \\ Events\end{tabular} & Nursing and physician notes, ECG reports, radiology reports, and discharge summaries. & notevents & 2.0 (0.52) & 0.5 (0.66) \\ \hline
Prescriptions & Medications ordered for a patient. & prescriptions & 4.0 (1.04) & 1.9 (2.51) \\ \hline

\end{longtable}

\subparagraph{}
The in-hospital mortality rate was 9.68\%, and the rate of length of stay in ICU $\geq$7 days was 19.73\%. The median age of subjects was 64 years with a median length of stay in ICU of 2.7 days and a median length of stay in hospital of 7.0 days. The male to female ratio was 56.36:43.64 (Table \ref{table:DataChars}). Table \ref{table:Dist_ICD9codes} displays the distribution of the primary International Classification of Diseases, 9th Edition (ICD-9) codes for patients. Diseases of the circulatory system (390-459, ICD-9 codes) as a primary diagnosis were observed for 34.11\% of hospital admissions), and 14.52\% had a primary diagnosis related to injury and poisoning (800-999, ICD-9 codes).

\begin{longtable}[H]{@{}ll@{}}
	\caption{Characteristics of the dataset} \tabularnewline
	\label{table:DataChars}
	
	Characteristics & Details \tabularnewline
	\hline

	Number of Subjects                   & 35348\tabularnewline
	Number of Admissions                 & 42818\tabularnewline
	Age, median(Q1, Q3)                  & 64 (48, 77)\tabularnewline
	Subject Gender(Female)               & 43.64\%\tabularnewline
	Subject Gender(Male)                 & 53.36\%\tabularnewline
	Admission Gender(Female)             & 43.91\%\tabularnewline
	Admission Gender(Male)               & 56.09\%\tabularnewline
	In-hospital Mortality                & 9.68\%\tabularnewline
	ICU stay(days), median(Q1, Q3)       & 2.7 (1.7. 5.6)\tabularnewline
	Hospital stay(days), median(Q1, Q3)  & 7.0 (4.0. 12.0)\tabularnewline
	ICU stay $\geq7$ Days                & 19.73\%\tabularnewline  
	\hline
\end{longtable}

\begin{longtable}[H]{@{}lrr@{}}	
	\caption{Distrubution of Primary ICD-9 codes for Patients} \tabularnewline
	\label{table:Dist_ICD9codes}
	
	Disease Distribution (ICD-9)                  &   count    & \% of Total \tabularnewline
	\hline
	Diseases of circulatory system (390–459)      &14607 &34.11\tabularnewline 
	Injury and poisoning (800–999)                &6219  &14.52\tabularnewline 
	Diseases of digestive system (520–579)        &3946  &9.22\tabularnewline 	
	Diseases of respiratory system (460–519)      &3801  &8.88\tabularnewline 	
	External causes of injury                     &3692  &8.62\tabularnewline  	
	Infectious and parasitic diseases (001-139)   &3367  &7.86\tabularnewline 	
	Neoplasms (140-239)                           &2617  &6.11\tabularnewline 	
	Endocrine, and metabolic diseases (240-279)   &1062	 &2.48\tabularnewline 	
	Diseases of genitourinary system (580–629)    &769   &1.80\tabularnewline 	
	Diseases of nervous system (320–389)          &628   &1.47\tabularnewline 
	Diseases of musculoskeletal system (710–739)  &488   &1.14\tabularnewline  
	ill-defined conditions (780–799)              &417   &0.97\tabularnewline 	
	Mental disorders (290–319)                    &351   &0.82\tabularnewline 	
	Congenital anomalies (740–759)                &244   &0.57\tabularnewline 	
	Conditions in the perinatal period (760–779)  &200   &0.47\tabularnewline 	
	Diseases of blood (280-289)                   &168   &0.39\tabularnewline 	
	Diseases of the skin (680–709)                &122   &0.28\tabularnewline 	
	Complications of pregnancy (630–679)          &120   &0.28\tabularnewline  
	\hline
\end{longtable}

\subsection{Data Engineering}
Tables in the MIMIC dataset were combined to create \textit{AllSources}: Chart Events, Input Events, Output Events, Lab Events, Microbiology Events, Procedure Events, Note Events, and Prescriptions (Table \ref{tab:table_datasources}). \textit{AllSources} were tokenized to prepare the data for training. Tokenization is the process of segmentation for generating tokens from data, which form the basic units for analysis.\cite{webster-kit-1992-tokenization} Features in each data source contained the following types of data: free-text data, code for recorded observation, measurement of the observation, and metadata about the measurement, e.g., unit of measure. For each data source, features containing free-text data were split on whitespace to create tokens, and other features were combined to create tokens. 

\begin{figure}[H]
\centering
\caption{Illustration of data tokenization.}
\label{fig:tokenization_process}
\includegraphics[width=0.85\textwidth]{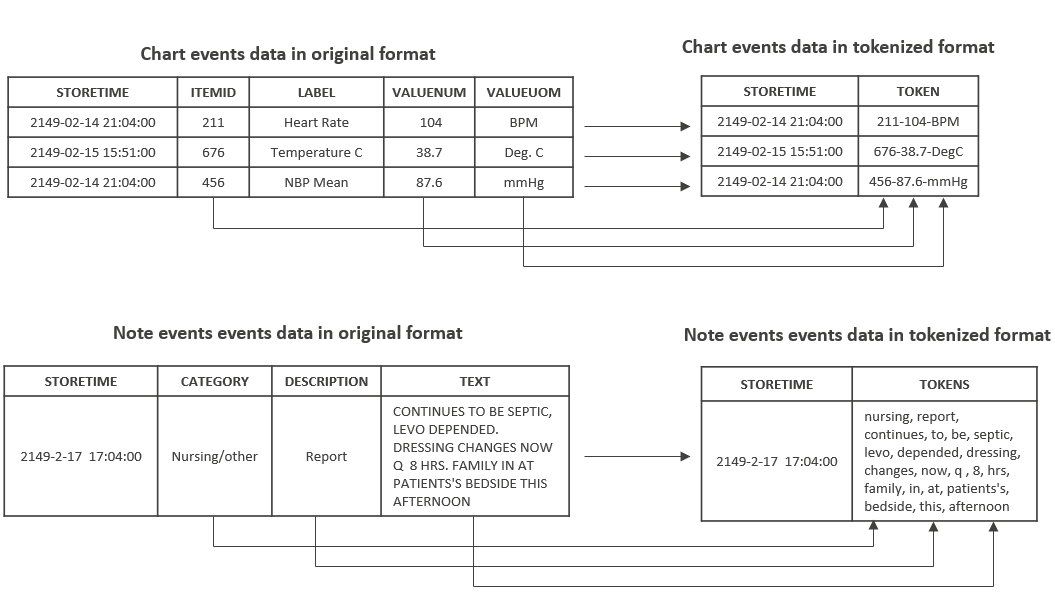}
\caption*{The features in chart events data such as ITEMID, VALUENUM, and VALUEUOM were combined to create tokens. LABEL shows the type of measurement, and ITEMID represents a unique code for each LABEL. VALUENUM contains the recorded value, and VALUEUOM shows the unit of measurement for a given value. Free-text data, i.e., TEXT in note events, was split on whitespace to create a sequence of tokens. CATEGORY describes the type of report, and DESCRIPTION shows a brief description of the notes. STORETIME indicates the time at which data was recorded in the database. Though not shown in the image, LABEL in chart events was also tokenized in a format similar to note events. Other sources of data, such as input events, output events, lab events, microbiology events, procedure events, and prescriptions, were tokenized in a similar format.}
\end{figure}

Figure \ref{fig:tokenization_process} illustrates the tokenization process with examples from Chart Events and Note Events. Chart Events contained features such as ITEMID, VALUE, and VALUEUOM. ITEMID describes an observation, VALUE is the measurement, and VALUEUOM is the unit of measure. These features were combined to create tokens, e.g., ‘211-104-BPM’, where ‘211’ is the unique code for ‘Heart Rate,’ ‘104’ is the value, and ‘BPM’ is the unit of measurement. Free-text features such as LABEL or physician comments or notes were split on white space into a sequence of tokens. For example, ‘Heart Rate’ would be split into two tokens ‘Heart’ and ‘Rate.’ All missing values were replaced with ‘NaN.’ Other sources of data, such as input events, output events, lab events, microbiology events, procedure events, and prescriptions, were tokenized in a similar format. This tokenization methodology is similar to the one used in a recent study by Rajkomar et al.\cite{Rajkomar2018}

\begin{figure}[H]
	\centering
		\caption{Illustration of the process to transform and merge tokenized data sources.}
		\label{fig:data_merge_process}	
		\includegraphics[width=0.85\textwidth]{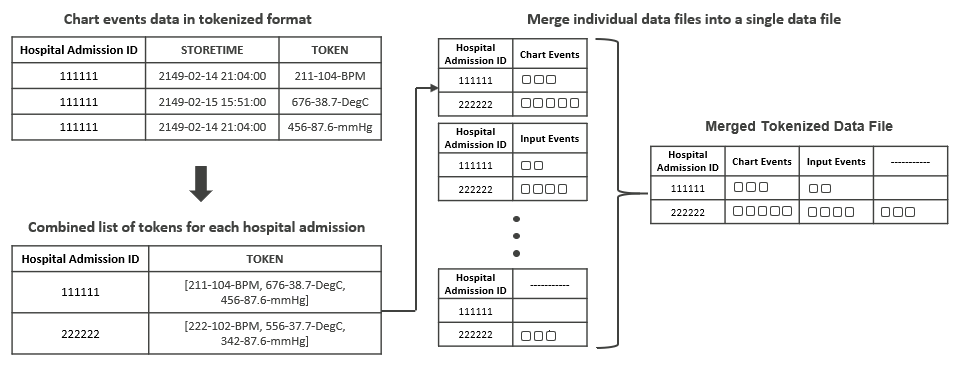}
		\caption*{STORETIME is the timestamp associated with the events. It represents the time at which data was recorded into the database. Tokens occurring for each hospital admission in Chart Events were combined into a single list of tokens. This process was repeated for each data source. Next, all data sources were merged on hospital admission to form a single dataset.}
\end{figure}

\begin{figure}[H]
	\centering
		\caption{Integer encoding of tokens.}
		\label{fig:integer_encoding}
		\includegraphics[width=0.85\textwidth]{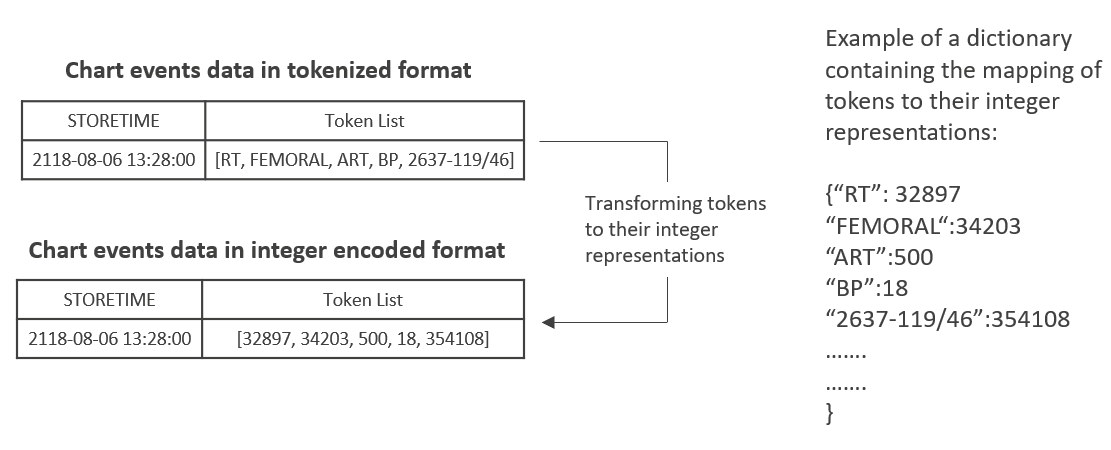}
		\caption*{The tokenized data from chart events was transformed into integer encoded data, where a unique integer represented each token. Tokens that occur only once were mapped to single out of vocabulary token.}
\end{figure}

Observations occurring in the first 24 hours of the ICU stay were extracted for each data source. For each data source, all tokens occurring for the same hospital admission were combined into a single list of tokens. Then all data sources were merged on hospital admission resulting in a single dataset (Figure \ref{fig:data_merge_process}). A vocabulary of unique tokens, a dictionary containing the token frequency, and a dictionary containing integer representations of tokens were created for each data source. Any token that occurred more than once was assigned its own integer representation (Figure \ref{fig:integer_encoding}), while tokens, which occurred only once, were mapped to a single out-of-vocabulary token.


Two timestamps, CHARTTIME and STORETIME, were available for each observation. CHARTTIME indicates the time at which the observation was made, and STORETIME shows the time at which the observation was manually inputted into the system or validated by clinical staff. This study relied on STORETIME as it was the closet proxy to when the data was recorded into the system. STORETIME was not available for LAB EVENTS, MICROBIOLOGY EVENTS, and PRESCRIPTION so, CHARTTIME or STARTDATE was used instead.

\subsection{Experiments and Model Training}
The transformed data comprised 35,348 patients and 42,818 hospital admission. All patients were randomly split into training and test set in 85:15 ratio. Then the training list was further divided into training and validation set in the ratio 85:15. These patient lists were used to create training, validation, and test dataset. This was done to ensure that the training, validation, and test dataset do not share hospital admissions from the same patient. The test dataset was used for final evaluation only, to prevent overfitting. The training dataset and validation dataset consisted of 30,963 and 5,431 hospital admissions, respectively. The test dataset comprised 6,424 hospital admissions. DLMs were constructed for two clinical outcomes, In-hospital mortality (IHM) and length of stay in ICU $\geq$ 7 days (LOS). Two models were created for each clinical outcome, resulting in a total of four models. For each outcome, the first model used only Chart Events, and the second model used \textit{AllSources} (Table \ref{tab:table_modeloverview}). The same training, validation, and test datasets were used to build all four models. The rate of mortality in the test dataset was 9.68\%, and the rate of length of stays $\geq$ 7 days was 20.03\%.

\begin{table}[H]
\centering
\caption{An overview of the models built, and data source used for training.}
\label{tab:table_modeloverview}
\begin{tabular}{lllr}
\hline
\multicolumn{1}{c}{Models} & \multicolumn{1}{c}{Outcomes} & \multicolumn{1}{c}{Data Source} & \multicolumn{1}{c}{\begin{tabular}[c]{@{}c@{}}Rate of outcome \\ (Test Dataset)\end{tabular}} \\ \hline
IHM-CS & In-Hospital Mortality (IHM) & Chart Events & 9.68\% \\
IHM-AS & In-Hospital Mortality (IHM) & \textit{AllSources} & 9.68\% \\
LOS-CS & Length of stay $\geq$ 7 days (LOS) & Chart Events & 20.03\% \\
LOS-AS & Length of stay $\geq$ 7 days (LOS) & \textit{AllSources} & 20.03\% \\ \hline
\end{tabular}
      \caption*{Models for predicting In-hospital mortality (IHM) and Length of stay in ICU $\geq$7 days (LOS) using only Chart Events are denoted as IHM-CS and LOS-CS, respectively. Models trained to predict IHM and LOS using all data sources are denoted as IHM-AS and LOS-AS, respectively. \textit{AllSources} comprises Chart Events, Input Events, Output Events, Lab Events, Microbiology Events, Procedure Events, Note Events, and Prescriptions.}
 
\end{table}

All models were trained to minimize Binary Cross-Entropy, using Adam optimizer. For each model, 120 trials were conducted to identify hyperparameters using Bayesian optimization, resulting in a total of 480 trials across all four models. The embedding dimension, rate of embedding dropout, number of dense layers, neurons per layer, rate of dropout for hidden layers, and learning rate were tuned as part of the hyperparameter tuning. The hyperparameter’s values that resulted in the least error on the validation dataset were selected as the best settings. Models trained using only Chart Events took roughly 26 hours to finish training, and models trained using all data sources took nearly 44 hours to complete training. Training all four models required a total of 140 hours.  

Python (Python Software Foundation) and TensorFlow were used for data engineering and model construction. Big Query was used for storing, analyze, and query the data. Big Query is a serverless auto-scaling cloud data warehouse on Google Cloud Platform (GCP). Models were trained using NVIDIA Tesla K80 GPU (Graphics Processing Unit). The CPU (Central Processing Unit) and memory resources were scaled as needed during the study. The hyperparameter training was performed using AI Platform on GCP. AI Platform recommends setting the number of trials not less than 10 times the number of hyperparameters. As 6 hyperparameters were tuned, 120 trials (twice the recommended number) were conducted for each model. The code used for conducting this study has been made publicly available at https://github.com/Rediminds/All-Data-Inclusive-Deep-Learning-Models-to-Predict-Critical-Events-in-MIMIC-III.

\subsection{Model Architecture}
\begin{figure}[H]
	\centering
		\caption{The architecture of the model trained using only chart events.}
		\label{fig:chartevents_arch}
		\includegraphics[width=\textwidth]{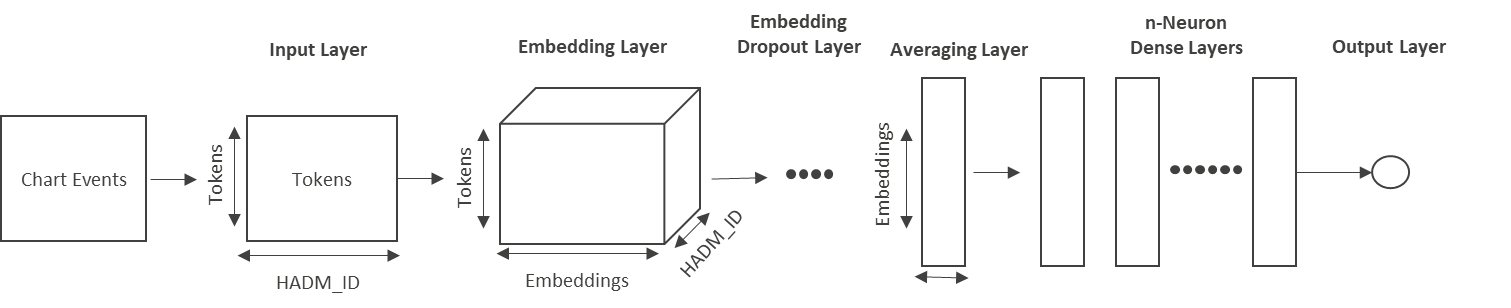}
		\caption*{This network consists of the following layers: an input layer, an embedding layer, an embedding dropout layer, an averaging layer, multiple dense hidden layers, and an output layer. The output layer was implemented with a sigmoid activation function. The network minimizes the loss function, i.e., binary cross-entropy, and uses Adam optimizer.}
\end{figure}

In this study, deep learning networks were created using the following layers: input layer, embedding layer, dropout layer, averaging layer, dense layer, and output layer.\cite{goodfellow2016deep,chollet2017deep,geron2017hands} Figure \ref{fig:chartevents_arch} illustrates the structure of the network trained using data from only Chart Events. The input to the network was a two-dimensional array. This array comprised of n (batch size) hospital admissions and a list of tokens for each hospital admission. The embedding layer mapped discrete tokens to a d-dimensional vector of real numbers. A spatial dropout layer was implemented to drop the embedding of tokens selected at random. Then a pooling layer combined the embeddings by averaging along the token axis. Averaged embeddings were passed to a multi-layered dense network with a dropout.\cite{srivastava2014dropout} The output layer comprised a single neuron with a sigmoid activation function.

\begin{figure}[H]
	\centering
		\caption{The architecture of the model using all events.}
		\label{fig:allevents_arch}
		\includegraphics[width=\textwidth]{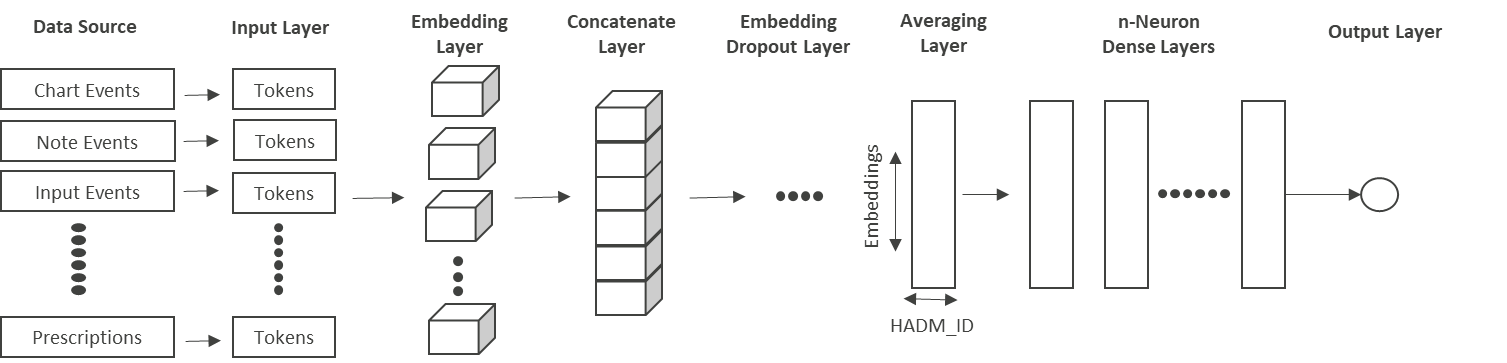}
		\caption*{This network was constructed using the following layers: input layers, embedding layers, a concatenation layer, an embedding dropout layer, an averaging layer, multiple dense layers, and an output layer. The output layer was implemented with a sigmoid activation function. The network minimized binary cross-entropy, the loss function, and used Adam optimizer.}
\end{figure}

The structure of the network trained using all data sources is shown in  figure \ref{fig:allevents_arch}.  The network consisted of multiple input layers. Each input layer was connected to a data source and accepted a two-dimensional array as input. Input array comprised of n (batch size) hospital admissions and a list of tokens for each hospital admission. Every input layer was connected to an embedding layer that mapped discrete tokens to a d-dimensional vector of real numbers. Embeddings from all data sources were merged using a concatenation layer. The merged embeddings were passed to a spatial dropout layer, which dropped the embedding of tokens selected at random. Then a pooling layer averaged the embeddings along the token axis. Averaged embeddings were passed to a dense network with dropout. The output layer comprised a single neuron with a sigmoid activation function.

\subsection{Model Evaluation and Statistical Analysis}
\label{subsection:model_eval}
The comparative performance of the models was assessed using area under receiver operating curve (AUC- ROC) and area under precision-recall curve (PR-AUC). In highly skewed datasets, PR-AUC is shown to be more informative in the evaluation of model performance.\cite{Davis:2006:RPR:1143844.1143874,Saito2015} Bootstrapping was used to generate confidence intervals for the scores.\cite{Rajkomar2018,2017arXiv170307771H,2016arXiv160805745C,Smith2008,Rajpurkar2017} The test dataset was resampled 10,000 times with replacement to create a bootstrapped population. This simulated population was used to generate a 95\% confidence interval (CI). A permutation test was performed to assess if the observed difference for both AUC-ROC and PR-AUC was significant. For each outcome, a model was determined to be statistically better performing if the 95\% CI excluded the point estimate of the other model, and if the permutation test resulted in a p-value less than 0.05. 

\section{Results}
\subparagraph{} 
Two models were constructed for each outcome, as described in subsection \ref{subsection:model_eval}, resulting in a total of four models (Table \ref{tab:table_modeloverview}). The models developed for predicting IHM using Chart Events and \textit{AllSources} are denoted as IHM-CS and IHM-AS, respectively. The models constructed for predicting LOS using Chart Events and \textit{AllSources} are denoted as LOS-CS and LOS-AS. From \textit{AllSources}, over 75 million events were formulated into tokens resulting in over 355 million tokens. The models for each outcome were evaluated using the test dataset comprising 6,424 hospital admissions. One hundred twenty trials were conducted for each model to identify the best hyperparameters. Hyperparameters used for training are presented in Table \ref{tab:table_hyperparamters}.

\begin{table}[H]
\centering
\caption{Hyperparameters used for training the models.}
\label{tab:table_hyperparamters}
\begin{tabular}{|p{2.5cm}|l|p{1.8cm}|p{1.8cm}|p{1.8cm}|p{1.8cm}|}
\hline
\multicolumn{1}{|l|}{\textbf{Hyperparameters}} & \multicolumn{1}{c|}{\textbf{Search space}} & \multicolumn{1}{c|}{\textbf{IHM-AS}} & \multicolumn{1}{c|}{\textbf{IHM-CS}} & \multicolumn{1}{c|}{\textbf{LOS-AS}} & \multicolumn{1}{c|}{\textbf{LOS-CS}} \\ \hline
Embedding dimension & {[}3, 15{]} & 3 & 10 & 5 & 14 \\ \hline
Embedding dropout rate & {[}0.0, 0.8{]} & \seqsplit{0.75590264936005447} & \seqsplit{0.0094835838251903226} & \seqsplit{0.69940654499820853} & \seqsplit{0.58825807853566} \\ \hline
Number of dense layers & \{1,2,3,4,5\} & 2 & 5 & 1 & 2 \\ \hline
Number of neurons per layer & \begin{tabular}[c]{@{}l@{}}\{16,32,64,\\ 128,256,512\}\end{tabular} & 64 & 128 & 32 & 64 \\ \hline
Dropout rate & {[}0.0,0.8{]} & \seqsplit{0.35016796955270379} & \seqsplit{0.072218911791479809} & \seqsplit{0.58363363148410985} & \seqsplit{0.64431304225324881} \\ \hline
Learning rate & {[}0.0001,0.1{]} & \seqsplit{0.0019374278041438577} & \seqsplit{0.00010326890647844106} & \seqsplit{0.0042757351885150414} & \seqsplit{0.0005435865071359092} \\ \hline
\end{tabular}
\caption*{Models for predicting In-hospital mortality (IHM) and Length of stay in ICU $\geq$ 7 days (LOS) using only Chart Events are denoted as IHM-CS and LOS-CS, respectively. Models trained to predict IHM and LOS using all data sources are denoted as IHM-AS and LOS-AS, respectively. Search space shows the range of values explored for each hyperparameter. The final hyperparameter settings are displayed in the column for each model}
\end{table}

\subsection{In-Hospital Mortality}
\begin{figure}[H]
	\centering
		\caption{Performance comparison of models IHM-AS and IHM-CS on the test dataset.}
		\label{fig:IHM_curves}
		\includegraphics[scale=0.75]{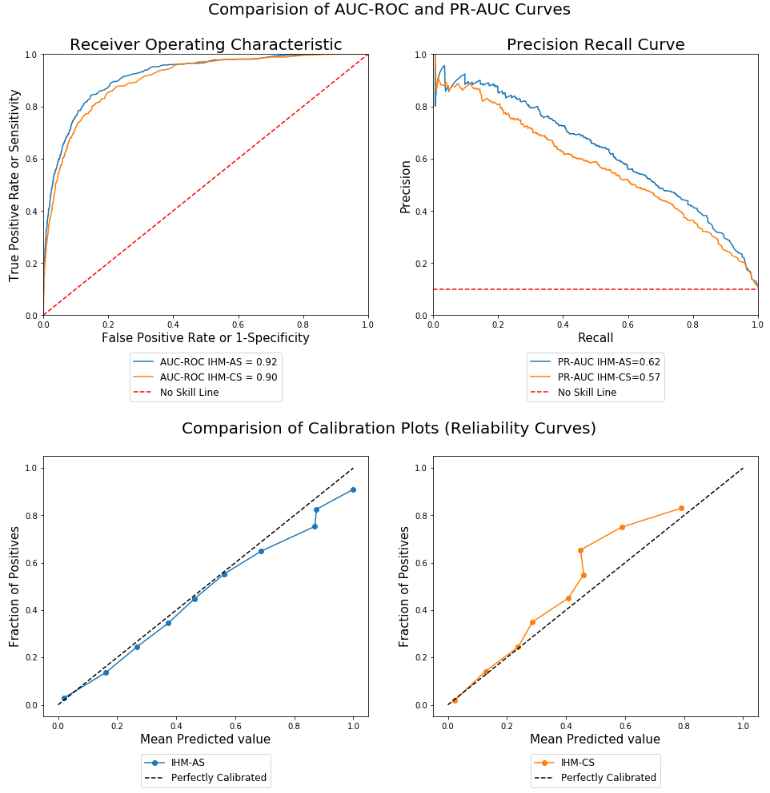}
		\caption*{The AUC-ROC curves and PR-AUC curves for IHM-AS and IHM-CS outperform the random classifier, indicated by the red dotted line. The black dotted line indicates a perfectly calibrated model. Both models appear to be well-calibrated as most points lie close to the perfect calibration line.}
\end{figure}

The F1 score for IHM-AS and IHM-CS was 0.517 and 0.518, respectively. For IHM-AS, precision and recall score was 0.725 and 0.402, respectively, and for IHM-CS, precision and recall score was 0.599 and 0.457, respectively. The F1 score, precision, and recall values reported above were observed at a threshold of 0.5.  Figure \ref{fig:IHM_curves}. illustrates the performance comparison of models IHM-AS and IHM-CS. IHM-AS and IHM-CS achieved an AUC-ROC of 0.918 and 0.903, respectively, and a PR-AUC of 0.625 and 0.570, respectively. Both models outperformed a random classifier, which is indicated by a red dotted line in Figure \ref{fig:IHM_curves}. Also, both models were observed to be well-calibrated as most of the points lie close to the perfect calibration line, indicated by the black dotted line in Figure \ref{fig:IHM_curves}. 

A bootstrapped population was simulated by resampling the test dataset 10,000 times with replacement. The AUC-ROC for model IHM-AS and IHM-CS was 0.9178 (95\%CI, 0.9062, 0.9285) and 0.9029 (95\%CI, 0.8902, 0.9144), respectively (Table \ref{tab:table_mortalityfit}). The PR-ROC for the IHM-AS and IHM-CS models was 0.6251 (95\%CI, 0.5828, 0.6651) and 0.5701 (95\%CI, 0.5275, 0.6115), respectively (Table \ref{tab:table_mortalityfit}). Observed AUC-ROC difference of 0.0149 between IHM-AS and IHM-CS was found to significant at 0.05 with a p-value of 0.0399, and the observed PR-AUC difference of 0.0549 was found to be significant at 0.05 with a p-value of 0.0366.

\begin{table}[H]
\centering
\caption{Model fit summary for in-hospital mortality (IHM).}
\label{tab:table_mortalityfit}
\begin{tabular}{ccc}
\text{Model} & \text{AUC-ROC (CLI-95\%)} & \text{PR-AUC (CLI-95\%)} \\ \hline
IHM-AS & 0.9178 (0.9062, 0.9285) & 0.6251 (0.5828, 0.6651) \\
IHM-CS & 0.9029 (0.8902, 0.9144) & 0.5701 (0.5275, 0.6115) \\ \hline
\end{tabular}
\caption*{The models developed for predicting IHM using Chart Events and \textit{AllSources} are denoted as IHM-CS and IHM-AS. Confidence intervals were generated by resampling the test dataset 10,000 times with replacement.}
\end{table}

\subsection{Length of Stay $\geq$ 7 days}
\begin{figure}[H]
	\centering
		\caption{Performance comparison of models LOS-AS and LOS-CS on the test dataset.}
		\label{fig:los_curves}
		\includegraphics[scale=0.75]{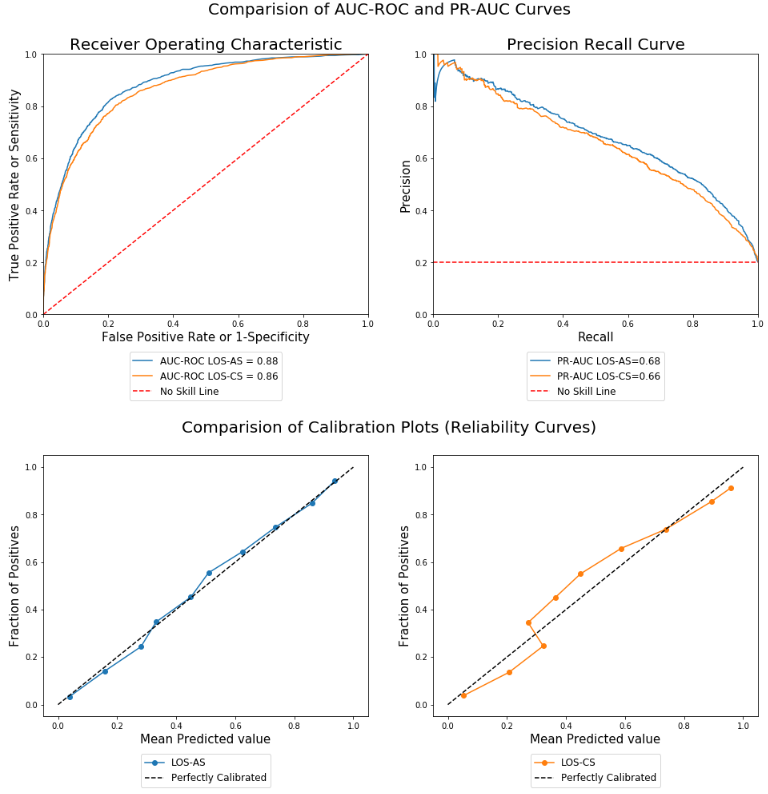}
		\caption*{The AUC-ROC curves and PR-AUC curves for LOS-AS and LOS-CS outperform the random classifier, indicated by the red dotted line. The black dotted line indicates a perfectly calibrated model. Both models appear to be well-calibrated as most points lie close to the perfect calibration line.}
\end{figure}

The F1 score for LOS-AS and LOS-CS was 0.605 and 0.603, respectively. For LOS-AS, precision and recall was 0.662 and 0.557, respectively, and for LOS-CS, precision and recall was 0.637 and 0.572, respectively. The F1 score, precision, and recall values reported above were observed at a threshold of 0.5.  Figure \ref{fig:los_curves} illustrates the performance comparison of models LOS-AS and LOS-CS. LOS-AS and LOS-CS achieved an AUC-ROC of 0.880 and 0.864, respectively, and a PR-AUC of 0.682 and 0.658, respectively. Both models outperformed a random classifier, which is indicated by a red dotted line in Figure \ref{fig:los_curves}. Also, both models were observed to be well-calibrated as most of the points lie close to the perfect calibration line, indicated by the black dotted line in Figure \ref{fig:los_curves}.

A bootstrapped population was simulated by resampling the test dataset 10,000 times with replacement. The AUC-ROC for LOS-AS and LOS-CS was 0.8806 (95\%CI, 0.8704, 0.8906) and 0.8642 (95\%CI, 0.8532, 0.8750), respectively (Table \ref{tab:table_losfit}). The PR-ROC for the LOS-AS and LOS-CS models was 0.6821 (95\%CI, 0.6552, 0.7074) and 0.6575 (95\%CI, 0.6311, 0.6829), respectively (Table \ref{tab:table_losfit}). Observed AUC-ROC difference of 0.016 between LOS-AS and LOS-CS was found to be significant at 0.05 with a p-value of 0.0147, and the observed PR-AUC difference of 0.0245 was found to be not significant at 0.05 with a p-value of 0.0983.

\begin{table}[H]
\centering
\caption{Model fit summary for length of stay$\geq$7 days (LOS).}
\label{tab:table_losfit}
\begin{tabular}{ccc}
\text{Model} & \text{AUC-ROC (CLI-95\%)} & \text{PR-AUC (CLI-95\%)} \\ \hline
LOS-AS & 0.8806 (0.8704, 0.8906) & 0.6821 (0.6552, 0.7074) \\
LOS-CS & 0.8642 (0.8532, 0.8750) & 0.6575 (0.6311, 0.6829) \\ \hline
\end{tabular}
\caption*{The models developed for predicting LOS using Chart Events and \textit{AllSources} are denoted as LOS-CS and LOS-AS. Confidence intervals were generated by resampling the test dataset 10,000 times with replacement.}
\end{table}

\section{Discussion}
Cost containment and optimal utilization of precious resources in an intensive care setting is a significant challenge before the critical care units across the world. Validated clinical tools and the predictive models could empower clinicians with better predictability and cost effectivity of decision making in a critical care setting. Currently, instruments such as APACHE, SAPS II, and SOFA are used discretely or in combination for risk stratification and identification of high-risk cohort.  They are limited by a lack of clear affirmation of their value in averting adverse patient outcomes. For the sake of convenience and ease of clinical application, these instruments are built to function with a limited number of variables such as 12 in APACHE II, and 17 in SAPS II.\cite{Naqvi2016} Despite the generation of vast amounts of data, much of the data remains unutilized. Recent advances in computing power, the resurgence of artificial intelligence, near-infinite cloud storage, and the availability of large databases have further facilitated building AI-based predictive models. 

In this study, the observed difference between IHM-AS and IHM-CS was found to be significant for AUC-ROC (0.0149, p=0.0399) and PR-AUC (0.0549, p=0.0366). The observed difference between LOS-AS and LOS-CS was noted to be significant for AUC-ROC (0.0164, p=0.0147) and not significant for PR-AUC (0.0245, p=0.0983). Therefore, it can be inferred that using more data sources in addition to Chart Events improved the performance of the model for predicting IHM and did not significantly improve the performance of the model for predicting LOS. It was also observed that learning from \textit{AllSources} did not decrease the performance of the model for predicting LOS. In this study, models for both outcomes were trained using the same dataset, eliminating the need for reliance on a few hand-selected variables. In a recent study, a similar methodology was used to create deep learning models trained to predict multiple clinical outcomes. They constructed a single dataset from multiple FHIR resources, greatly reducing the effort spent on data preparation and variable selection for each outcome.\cite{Rajkomar2018} 

Harutyunyan et al., Yu et a., and Si et al. constructed multitask neural networks to predict in-hospital mortality and length of stay using only few physiologic variables\cite{2017arXiv170307771H, yu2019using} or caregiver notes\cite{si2019deep}. Harutyunyan et al. and Yu et al. resampled the data at regular intervals, and Si et al. used only clinical notes, thus discarding large amounts of patient data. Unlike the works discussed above, in this study, the models for predicting IHM and LOS were trained using the first 24 hours of ICU data from \textit{AllSources}, and Chart Events without resampling time series data at regular intervals. Also, models in this study did not rely on hand-selected features and leveraged the same dataset for predicting both outcomes. Davoodi et al. trained a Depp Rule-Based Fuzzy System, where they investigated rule generation using clustering on mixed attributes. In contrast to models constructed in this study, they leveraged only a few physiologic variables and resampled the data at regular intervals over a window of 48 hours.\cite{DAVOODI201848} Boag et al. leveraged only the first 24 notes to predict in-hospital mortality.\cite{boag2018s} Jin et al. proposed a multimodal neural network trained using time series variables and unstructured clinical notes from the first 48 hours of ICU stays. They used clinical notes, and few physiological variables, discretized every 2 hours.\cite{jin2018improving} This also resulted in discarding vast amounts of patient data. Weissman et al. evaluated the performance of multiple machine learning models for predicting a composite outcome, i.e., in-hospital mortality or length of stay $\geq$7 days.\cite{weissman2018inclusion} The clinical utility of a composite score as formulated in their study is limited as different protocols would be required to treat patients with high mortality risk in comparison to patients with a high probability of extended hospital stay.

The dataset in this study, comprised over 75 million events from over 9000 items recorded across \textit{AllSources}. The models for predicting IHM using \textit{AllSources}, i.e., IHM-AS achieved an AUC-ROC of 0.9178 (95\%CI, 0.9062, 0.9285) and PR-AUC of 0.6251 (95\%CI, 0.5828, 0.6651). The models trained using \textit{AllSources} for predicting LOS, i.e., LOS-AS achieved an AUC-ROC of 0.8806 (95\%CI, 0.8704, 0.8906) and PR-AUC of 0.6821 (95\%CI, 0.6552, 0.7074). IHM-AS achieved higher AUC-ROC and PR-AUC scores than the models proposed by Harutyunyan et al., Davoodi et al., and Jin et al., which used a window of 48 hours. It also outperformed models proposed by Yu et al\cite{2017arXiv170307771H,DAVOODI201848,jin2018improving}. and Boag et al\cite{boag2018s}. The models constructed by Si et al. achieved higher performance for both in-hospital mortality and length of stay task, but the time window of the dataset used in their study was not provided\cite{si2019deep}. Harutyunyan et al. defined length of stay as a regression task\cite{2017arXiv170307771H}, and Boag et al. defined multiple length of stay classification tasks where all tasks were shorter than seven days\cite{boag2018s}.

The contributions of this study are three-fold. Firstly, this study reports a data engineering process to combine patient data comprising a variety of sources from an open-source dataset for ICU patients. Secondly, this study does not sample the dataset at regular time intervals, nor it relies on hand-picked features. Instead, it leverages data recorded at time intervals as available in the dataset. Thirdly, it reports a comparison of deep learning models that learned from all data sources against models constructed using only chart data.

In this study, there are some constraints and limitations. First, the models trained in this study used data specific to MIMIC-III dataset (Beth Israel Deaconess Medical Center). Hence, applying these models to data from other hospitals will require extensive data harmonization. Second, free-text data was tokenized on whitespace to create unigrams. Further studies are required to assess if adopting different methodologies to process the free-text data can help improve the performance of the model.\cite{webster-kit-1992-tokenization,habert1998towards,baeza1999modern} Third, the tokens for each hospital admission were combined into a single list ignoring the temporal nature of the data. While maintaining this structure of the events may allow us to take advantage of the temporal nature of the data, this is a subject of future research. Fourth, the deep learning models are often referred to as black-box models as the logic of output is cryptic to both clinicians and engineers who developed them. Though efforts have been made to interpret deep learning models, it remains an unfinished task.\cite{chakraborty2017interpretability,montavon2018methods} Fifth, though our study demonstrates the usage of more relevant data from a patient’s EHR can significantly improve the performance of predictive models, the impact of each data source such as caregiver notes requires further study. Sixth, we did not have access to the source medical center to deploy and validate the models in a prospective setting.

The objective of this study was not to rehash the superiority of machine learning techniques over the conventional statistical models but to demonstrate the importance of using complete patient data. A judicious application of a multitude of data engineering techniques is required to improve the performance of predictive models. These models, if created in collaboration with the clinicians, could develop into reliable clinical decision support tools.\cite{Deo2016} Ongoing interaction and synergy between machine and the clinicians make a formidable combination of human intelligence and artificial intelligence, leading towards augmented intelligence for the benefit of the patients.

\section{Conclusion}
In this study, an open-source critical care dataset (MIMIC-III) was used to construct models for predicting in-hospital mortality, and length of stay $\geq$7 days. Two models were constructed for each outcome, the first model was trained using all relevant data available (\textit{AllSources}), and the second model was trained using only one data source (Chart Events). It was observed that it is possible to predict in-hospital mortality with much better confidence and higher reliability from models built with \textit{AllSources}. The model for predicting length of stay $\geq$7 days using \textit{AllSources} performed marginally better, though not statistically significant, than the model built using only Chart Events. This study has three significant contributions. First, it reports a process to combine data from a variety of sources on an open-source dataset. Second, the models developed in this study did not rely on hand-selected features. Third, it reports the effectiveness of models trained using all data sources against models trained using only one data source. With this approach, we observed promising model performance from only 42,818 admissions. With a higher volume of patients, more variety of data sources, and deep learning, one may achieve much better model performance. In this study, we observed the potential to build the next generation of clinical practice tools that can augment the efficiency and effectiveness of caregivers in critical care. However, it is only possible when the data is digitized and ready to be used for model development.

\section{Aknowledgement}
Funding: We are grateful to RediMinds Inc. for funding this work. This publication only reflects the authors views. Funding agency is not liable for any use that may be made of the information contained herein.

\section{Declarations of interest}
Declarations of interest: none.



\section{References}
\bibliographystyle{elsarticle-num} 
\bibliography{MIMIC3_citaions}





\end{document}